\documentclass[letterpaper, 10 pt, conference]{ieeeconf}  %

\IEEEoverridecommandlockouts                              %

\usepackage{hyperref}
\usepackage{graphicx}
\usepackage{amsmath,amssymb,amsfonts}
\usepackage{algorithm, algpseudocode}
\usepackage{subcaption}
\usepackage{booktabs}
\usepackage{wrapfig}
\usepackage{relsize}  %
\usepackage{multirow}
\usepackage{caption}
\usepackage{stfloats}

\usepackage[
    style=ieee,
    natbib=true,
    citestyle=numeric-comp,
    doi=false,
    isbn=false,
    url=false]{biblatex} 
\newcommand{\sysName}{Refinery}

\title{\LARGE \bf
Refinery: Active Fine-tuning and Deployment-time Optimization \\ for Contact-Rich Policies
}

\author{ \authorblockN{Bingjie Tang$^{1}$, Iretiayo Akinola$^{2}$, Jie Xu$^{2}$, Bowen Wen$^{2}$, Dieter Fox$^{3}$, Gaurav S. Sukhatme$^{1}$, \\
Fabio Ramos$^{2,4}$, Abhishek Gupta$^{3}$, Yashraj Narang$^{2}$}
\authorblockA{ 
$^1$University of Southern California, $^2$NVIDIA Corporation, $^3$University of Washington, $^4$University of Sydney \\
}
}

\begin{document}

\setcounter{figure}{1}
\makeatletter
\let\@oldmaketitle\@maketitle%
\renewcommand{\@maketitle}{
   \@oldmaketitle%
   \begin{center}
    \centering      
    \includegraphics[width=\linewidth]{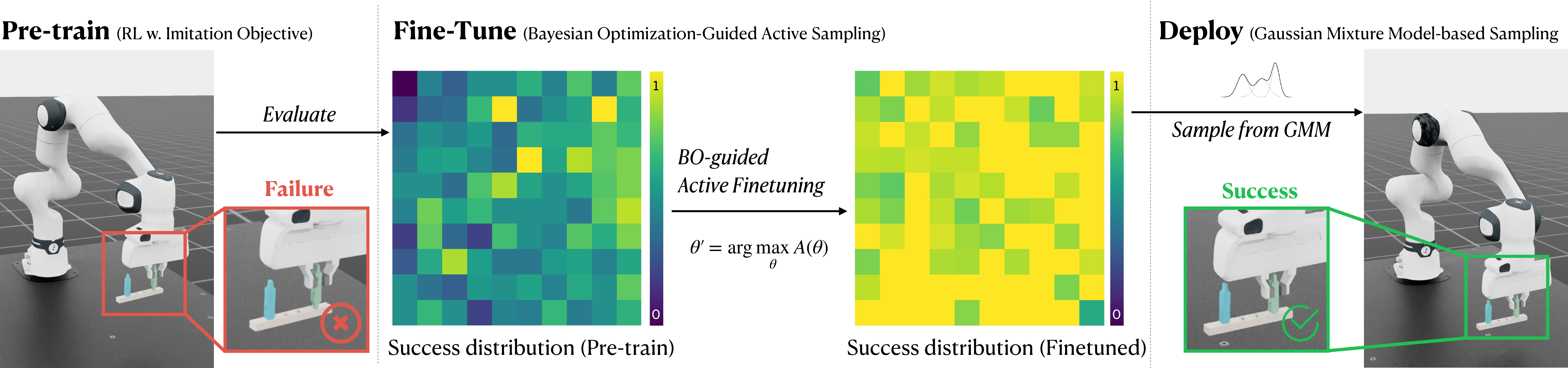}
  \end{center}
  \footnotesize{Fig.~\thefigure:\label{fig:teaser}~ \textbf{Method Overview}. We pretrain RL policies for assembly and evaluate them over initial conditions. (Here, a success map is shown over initial x- and y-coordinates). We leverage Bayesian Optimization-guided fine-tuning to improve policy performance. We then propose Gaussian Mixture Model-based sampling during deployment to select initializations that maximize execution success.
  }
}
\makeatother

\maketitle
\thispagestyle{empty}
\pagestyle{empty}

\begin{abstract}

Simulation-based learning has enabled policies for precise, contact-rich tasks (e.g., robotic assembly) to reach high success rates ($\sim\!80\%$) under high levels of observation noise and control error. Although such performance may be sufficient for research applications, it falls short of industry standards and makes policy chaining exceptionally brittle. 
A key limitation is the high variance in individual policy performance across diverse initial conditions.
We introduce {\normalfont Refinery}, an effective framework that bridges this performance gap, robustifying policy performance across initial conditions. 
We propose Bayesian Optimization-guided fine-tuning to improve individual policies, and Gaussian Mixture Model-based sampling during deployment to select initializations that maximize execution success. 
Using {\normalfont Refinery}, we improve mean success rates by $10.98\%$ over state-of-the-art methods in simulation-based learning for robotic assembly, reaching $91.51\%$ in simulation and comparable performance in the real world. 
Furthermore, we demonstrate that these fine-tuned policies can be chained to accomplish long-horizon, multi-part assembly—successfully assembling up to 8 parts without requiring explicit multi-step training. See our \href{https://refinery-2025.github.io/refinery/}{project website} for more details.

\end{abstract}

\section{INTRODUCTION}

Robotic assembly is a longstanding challenge, requiring high-precision, high-accuracy contact-rich manipulation, often over long horizons. Simulation and sim-to-real transfer have emerged as powerful strategies for tackling these challenges. Recent advances in simulation-based learning have enabled the development of robust assembly policies that achieve strong performance in both simulated and real environments \cite{zhang2023bridging, zhang2023efficient, tang2023industreal, noseworthy2025forge}. Notably, these methods have achieved success rates of up to 80\% on challenging benchmarks involving 2-part assemblies
\cite{tang2024automate, guo2025srsa}.
Despite this progress, these results fall short of the requirements in industrial settings, where success rates of 95\% or higher are typically expected, and 90\% is considered a minimum threshold \cite{gencer2024first}.
These standards are even more critical in multi-part assembly, as failure rates compound across sequential stages.

This leads to a natural question: how do the current state-of-the-art assembly policies fail? Our key observation is that a policy with a given average success rate in simulation does not fail with constant probability across all initial conditions (see Figure 1.); rather, it succeeds from certain initial states and fails from others, with high spatial variance. This observation suggests that minimizing the success rate variance among all initial conditions during learning and prioritizing high-performing ones during deployment could substantially improve overall task performance, potentially enabling long-horizon multi-part assembly as well.

Motivated by this insight, we introduce \textbf{Refinery}, a framework designed to enhance the fine-tuning and deployment of learned contact-rich policies. Our core idea is simple but effective: (1) actively identify and fine-tune policies on initial states with high uncertainty, and (2) prioritize high-success initializations during deployment. This approach significantly improves both individual and sequence-level success, moving research-grade policies closer to industry-grade reliability. 

Specifically, our main contributions are: (1) \textbf{Active Sampling Method}: We introduce a Bayesian Optimization (BO)-guided fine-tuning strategy that significantly improves policy performance by focusing on high-uncertainty initial states; (2) \textbf{Deployment-time Optimization Method}: We propose a Gaussian Mixture Model (GMM)-based sampling strategy at deployment time that maximizes the success rate of already-trained or fine-tuned policies by prioritizing effective initial conditions; (3) \textbf{2-Part Assembly Results}: Using our approach, we achieve a 10.98\% improvement in success rate on the simulated benchmark of 100 two-part assemblies from \cite{tang2024automate}, and we demonstrate zero-shot sim-to-real transfer with 97\% success over 10 assemblies—exceeding prior state-of-the-art; (4) \textbf{Multi-part Assembly Results}: We provide 5 multi-part assemblies (derived from \cite{tian2023asap}) and corresponding simulation environments, and train assembly policies to solve these tasks. 
To our knowledge, this is the first work to demonstrate zero-shot sim-to-real for multiple long-horizon multi-part assemblies.

\section{RELATED WORK}
\label{sec:related-work}
\subsection{Robotic Assembly}

Robotic assembly has historically been addressed with analytical methods 
\cite{whitney_mechanical_2004, mason2001mechanics, drake1978using, lozano1984automatic, xia2006dynamic}. 
Recent efforts have shifted toward learning-based approaches for greater adaptability and generalization.
Model-based methods like GPS and iLQG offer sample efficiency but struggle with discontinuous contact dynamics \cite{thomas_learning_2018, luo_reinforcement_2019, spector_deep_2020}. 
Model-free, off-policy reinforcement learning (RL) methods (e.g., DQN, SAC, DDPG) improve sample reuse but suffer from convergence instability \cite{zhang_learning_2021, beltran-hernandez_variable_2020, luo2021robust}. 
On-policy RL methods (e.g., PPO, TRPO, A3C) are more stable but sample-inefficient \cite{hebecker_towards_2021, shao_learning_2020}.
Several works leverage demonstrations to improve data efficiency using residual learning, guided DDPG, DDPGfD, or offline RL \cite{johannink2019residual, fan_learning_2019, vecerik_leveraging_2018, zhao2022offline}. 
However, collecting reliable demonstrations is often costly and time-consuming. 
Other non-RL methods include self-supervised learning from multimodal inputs or video-based imitation \cite{spector2021insertionnet, wen2022you}.
While many methods achieve high success rates, they often rely on human input, long training times, or real-world resets, limiting scalability. 

Over the past few years, advances in simulation accuracy and speed have enabled large-scale policy training for robotics, including contact-rich tasks like assembly \cite{isaaclab2025, narang2022factory, Genesis, todorov2012mujoco}. 
There are few directly-comparable works for simulation-based assembly policies \cite{tang2024automate, guo2025srsa, jiang2024transic, ankile2024juicer}.
Most focus on two-part assemblies or rely on human demonstrations, whereas our work leverages active sampling for diverse two-part and multi-part assembly tasks.

\subsection{Improving Subtask Policies for Long-Horizon Tasks}
Multi-part assembly can be decomposed into a sequence of two-part assembly subtasks, enabling multi-step policy chaining. Prior works address such chaining by adapting each sub-policy to the terminal states of its predecessor \cite{konidaris2009skill, bagaria2019option, clegg2018learning, lee2021adversarial}. Extensions of this idea include backpropagation through the chain using goal regression in symbolic planning \cite{kaelbling2011hierarchical, kaelbling2016pre} and bidirectional fine-tuning for long-horizon tasks \cite{chen2023sequential, chen2024scar}. These approaches are most effective when terminal state distributions are broad. However, in tight-tolerance assembly, transitions between subtasks are narrowly distributed, making such chaining brittle and sequential fine-tuning costly.

Instead, our approach improves
individual policy performance without explicitly optimizing inter-policy transitions. While on-policy RL suffers from high-variance gradient estimates due to random sampling, we mitigate this by leveraging active sampling during fine-tuning. Unlike exploration-focused strategies (e.g., entropy-regularized RL \cite{haarnoja2018soft}, GPS in high-reward regions \cite{levine2013guided}, or Bayesian optimization for sample efficiency \cite{muller2021local, jones1998efficient}), our objective is to directly maximize task success.

\section{Problem Description}
\label{sec:problem-description}

\textbf{Goal:} Our goal is to improve the overall performance of sequences of policies that solve challenging, long-horizon, contact-rich assembly tasks. Policy performance is highly sensitive to task initialization, as the tasks require high precision and accuracy. Hence, we study the problem of identifying and expanding the subsets of initial states for each individual policy that maximize the overall success rate for the sequence.  

\textbf{Assumptions:} We make the following assumptions: (1) The assembly sequence is given, as is typical for industrial assembly workflows. (2) A CAD or mesh file is available for each part, as is typical for industrial assets. (3) All parts have size and initial pose such that at least one grasp is feasible and sufficient to allow subsequent insertion.

\textbf{Problem Setup:} Given a multi-part assembly problem consisting of $N$ parts, the task is decomposed into $N - 1$ sequential stages, where at each stage $i \in {1, ..., N-1}$, a new part (\textit{plug}) is inserted into the current partially-assembled structure. We first train a specialist policy $\pi_i$ for each stage using the approach from \cite{tang2024automate}, which uses on-policy RL with an imitation objective. We then apply the proposed fine-tuning and deployment-time optimization for each policy. \footnote{We use the same hyperparameters, randomization ranges, observation noise levels and network structures for training policies with PPO as \cite{tang2024automate}.}

\textbf{Definitions:} Each stage is formulated as a Markov decision process (MDP), where the agent is a simulated robot operating in an environment containing the parts to be assembled. We define a state space $\mathcal{S}$, observation space $\mathcal{O}$, and action space $\mathcal{A}$. 
The observation space includes robot arm joint angles, end-effector pose, goal pose, and the delta between the current and goal poses, and the robot action at each step is the delta between the current and next end-effector pose. Our state-transition dynamics is defined by $\mathcal{T}: \mathcal{S} \times \mathcal{A} \rightarrow \mathcal{S}$, governed by rigid-body dynamics in our simulator. We define a randomized initial state distribution $\rho_0$, a reward function $R: \mathcal{S} \rightarrow \mathbb{R}$, and a discount factor $\gamma \in (0, 1]$. The episode consists of $N$ timesteps, and the horizon is $T$, where $T \leq N$. The return $G$ is defined as
\begin{equation}
    G(T) = \mathbb{E}_{\pi} \big[ \Sigma^{T-1}_{t=0}\gamma^t R(s_t) \big],
\end{equation}
which represents the expected sum of discounted rewards over the horizon. The objective is to train a policy $\pi : \mathcal{O} \rightarrow \mathbb{P}(\mathcal{A})$ that maximizes this return.

For each stage $i$, we define a task-success evaluation function $J_i(\theta)$, where $\theta \in \mathbb{R}^3$ is the $xyz$ position of the part relative to its goal and $J_i(\theta) \in [0,1]$ represents the success probability of executing policy $\pi_i$ starting from $\theta$.

\begin{figure*}
    \centering
    \includegraphics[width=\linewidth]{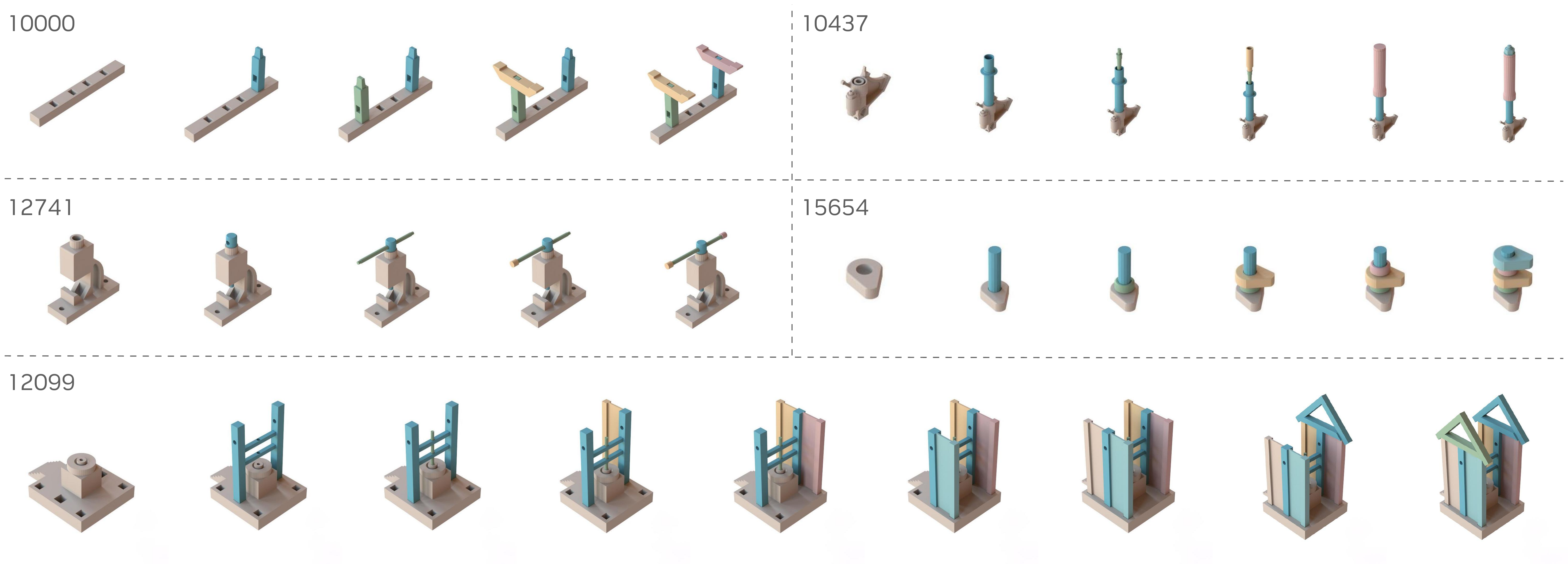}
    \caption{\textbf{Multi-part Assembly Dataset.} We provide 5 multi-part assemblies derived from \cite{tian2023asap}.}
    \label{fig:multipart_dataset}
\end{figure*}

\textbf{Dataset:} We created a geometrically-diverse dataset of 5 multi-part assemblies based on  \cite{tian2023asap} as shown in \autoref{fig:multipart_dataset}. 
We design our selecting criteria to ensure sufficient complexity and compatibility with our setup. 
Specifically, each assembly must consist of 5-9 mechanical components and require 4-8 discrete insertion-based steps to complete. 
All components must be graspable using a parallel-jaw gripper. 
Furthermore, the complete assembly must be executable in a sequential manner by a single arm, such that each part can be added atop the previously assembled parts placed in a static fixture.
For each assembly, we manually define a valid assembly sequence and designate the first component in this sequence as the \textit{base}.

For selected assemblies, we apply a multi-step procedure that involves (1) scaling all component meshes to fit within a 10 $cm^3$ bounding box; (2) reorienting the assembly such that the \textit{base} component is aligned with the z-axis in an upright configuration; (3) translating the assembly to ensure the bottom surface of the \textit{base} is coplanar with the global origin; and (4) applying a depenetration step to enforce a 0.5 $mm$ clearance between all parts. 
The resulting dataset is fully compatible with simulators that enforce non-penetration constraints \cite{isaaclab2025} and is 3D-printable in the real world.

\section{Method}
\label{sec:active-sampling}

We present \textbf{{\sysName}}, a method that leverages the policy success rate distribution as a prior to achieve two main objectives. First, it enables efficient sampling of high-uncertainty regions in the task initialization space during policy fine-tuning, aiming to \textit{improve every individual policy performance} (\autoref{subsec:method-finetuning}). Second, at deployment time, it prioritizes promising initialization conditions to \textit{maximize the overall success probability} of completing the full assembly task (\autoref{subsec:method-deployment}).

\subsection{Bayesian Optimization-Guided Fine-Tuning} 
\label{subsec:method-finetuning}
Given a sequence of policies $\pi_i$ for each stage $i$ of a multi-part assembly, our goal is to improve the overall sequence success rate by enhancing the robustness of individual policies through active fine-tuning.
Before fine-tuning, we first train a specialist policy $\pi_i$ for each stage with RL \cite{tang2024automate}.
During fine-tuning, we aim to actively identify and prioritize task initial states with high uncertainty of success, in contrast with the uniform random sampling when training from scratch. However, the underlying success-rate distribution across initial states is typically unknown, non-differentiable, non-convex, and expensive to evaluate exhaustively. Moreover, this distribution evolves as policy learning progresses.
To address these challenges, we adopt Bayesian Optimization (BO), a sample-efficient and gradient-free strategy for optimizing black-box functions. 
BO is particularly well-suited for our setting, as it enables informed exploration of the initialization space without requiring exhaustive evaluation.

In our formulation, for each stage $i$ of a multi-part assembly, we model the success rate $\hat{J}_i$ of policy $\pi_i$ as a function of $\theta$, the initial position of the \textit{plug} (i.e., the part to be inserted) using a Gaussian Process (GP):
\begin{equation}
    \hat{J}_i(\theta) \sim \mathcal{GP}(\mu(\theta), k(\theta, \theta'))
\end{equation}
where $\mu(\theta)$ is the mean function, and $k(\theta, \theta')$ is the covariance function that captures correlations between different initial states.
To model $\hat{J}_i$, we execute the policy $\pi_i$ from different randomly-sampled $\theta$ in simulation. \footnote{Sampling is parallelized and conducted entirely in simulation: executing 1000 rollouts takes only seconds on a single GPU.} 

After evaluation, BO then proposes the next sample $\theta'$ for fine-tuning by optimizing an acquisition function:
\begin{equation}
\label{eq:method-finetune-sample}
    \theta' = \arg \max_\theta A(\theta)
\end{equation}
where $A(\theta)$ is the selected acquisition function.

As described in Algorithm \ref{algo:finetune}, this process is repeated iteratively to guide fine-tuning toward initializations most likely to improve individual policy robustness. 
In each iteration, we re-evaluate the current policy and update $\hat{J}_i$; this updated distribution accounts for policy shifts and guides BO-based fine-tuning. We continue this process until each policy reaches convergence, defined as $<\!5\%$ variation in success rate over 5 consecutive epochs during evaluation.

\begin{algorithm}
\footnotesize
\caption{\small Bayesian Optimization (BO)-Guided Fine-Tuning}
\begin{algorithmic}[1]
\State \textbf{Input:} Assembly steps $\{A_1, A_2, ..., A_N\}$
\For{each assembly step $A_i$}
    \State \textbf{Train} specialist policy $\pi_{i}$ using \cite{tang2024automate}
    \While{fine-tuning not converged}
        \State \textbf{Evaluate} success rate distribution $J_i(\theta)$ at different plug initializations $\theta$
        \State \textbf{Use Active Sampling} (\autoref{eq:method-finetune-sample}) to propose samples $\theta'$
        \State \textbf{Fine-tune} each policy by initializing at proposed $\theta'$
    \EndWhile
\EndFor
\end{algorithmic}
\label{algo:finetune}
\end{algorithm}

In this work, we explored three acquisition functions:

1) \textbf{Upper Confidence Bound (UCB)}, which encourages exploration by selecting samples with high uncertainty:
\begin{equation*}
    A_{UCB}=\mu(\theta)+\beta\sigma(\theta)
\end{equation*} 
where  is a hyperparameter that controls the trade-off between exploitation and exploration, and $\mu(\theta)$ and $\sigma(\theta)$ are the predicted mean and standard deviation from the GP. \footnote{We evaluated several $\beta$ values and empirically choose $\beta=2.0$ in our experiments; however, our approach is not particularly sensitive to $\beta$.}

2) \textbf{Probability of Improvement (PI)}, which selects samples based on the likelihood of exceeding the best observed success rate $J^+$:
\begin{equation*}
    A_{PI}(\theta)=\Phi \left( \frac{\mu(\theta)-J^+}{\sigma(\theta)} \right)
\end{equation*}
where $J^+=\max J_i(\theta)$ is the best observed success rate, and $\Phi(\cdot)$ is the cumulative distribution function (CDF) of the standard normal distribution.

3) \textbf{Expected Improvement (EI)}, which selects the samples that are expected to improve over the best observed success rate $J^+$: 
$$A_{EI}(\theta)=(\mu(\theta)-J^+)\Phi(Z)+\sigma(\theta)\phi(Z)$$
where $Z=\frac{\mu(\theta)-J^+}{\sigma(\theta)}$ is the standardized improvement, and $\Phi(Z)$ and $\phi(Z)$ are the CDF and the probability density function of the standard normal distribution.

\subsection{GMM Sampling for Deployment-Time Optimization} 
\label{subsec:method-deployment}
While policy fine-tuning focuses on improving each policy's performance across different plug initializations, the focus at \textit{deployment} time is on identifying high-performing initial positions to maximize success rates.
More specifically, we want to sample plug initializations with high probability of execution success given the success rate distribution approximated from evaluation rollouts. 

To model the success distribution, we first collect a dataset by uniformly sampling $( N=1000 )$ initial plug poses $( \{x_i\}_{i=1}^N )$ from the predefined initialization space $( \mathcal{X} )$ and executing the fine-tuned policy $(\pi)$ from each initialization \footnote{Empirically, we found $N=1000$ rollouts to enable stable mixture fitting without overfitting.}.
Note that $x_i$ includes both initial plug position $\theta_i$ and orientation.
For each rollout, we record the binary success outcome $( y_i \in \{0,1\} )$. 
We then extract the subset $ \mathcal{D}_{\text{succ}} = \{x_i \mid y_i = 1\} $ from $( \mathcal{D} = \{(x_i, y_i)\}_{i=1}^N )$ corresponding to successful executions and fit a Gaussian Mixture Model (GMM) to this set to estimate the underlying distribution. 
At deployment time, we sample $( M = 1000 )$ candidate initializations $( \{\tilde{x}_j\}_{j=1}^M )$ from the GMM and compute the GMM probability density $( p(\tilde{x}_j) )$ for each sample. The final initialization is selected as $( x^* = \arg\max_j p(\tilde{x}_j) )$, i.e., the sample with the highest probability density under the GMM. 
This strategy enables efficient initialization selection by leveraging learned success distributions, avoiding brute-force search; a detailed description of the procedure is in Algorithm \ref{alg:gmm-sampling}.
\begin{algorithm}[H]
\footnotesize
\caption{\small Deployment-time Optimization via GMM Sampling}
\label{alg:gmm-sampling}
\begin{algorithmic}[1]
\State \textit{// Offline Phase: Data Collection and GMM Fitting}
\State \textbf{Sample} $N$ plug initializations $\{x_i\}_{i=1}^N \sim \mathcal{U}(\mathcal{X})$
\State \textbf{Execute} $\pi$ for each $x_i$, record binary success $y_i \in \{0, 1\}$, and collect a dataset of successful samples: $\mathcal{D}_{\text{succ}} = \{x_i \mid y_i = 1\}$
\State \textbf{Fit} a Gaussian Mixture Model with $K$ components to $\mathcal{D}_{\text{succ}}$:
\[
p_{\text{GMM}}(x) = \sum_{k=1}^{K} \xi_k \, \mathcal{N}(x \mid \mu_k, \Sigma_k)
\]
\Statex
\State \textit{// Online Phase: Deployment-time Plug Initialization Selection}
\State \textbf{Sample} $M$ candidates from the GMM: $\{x_j'\}_{j=1}^M \sim p_{\text{GMM}}(x)$
\State \textbf{Compute} GMM probability density $s_j = p_{\text{GMM}}(x_j')$ for each $x_j'$
\State \textbf{Select} $x^* = \arg\max_{x_j'} s_j$ and initialize the plug at $x^*$
\State \textbf{Execute} policy $\pi$
\end{algorithmic}
\end{algorithm}

\section{Experimental Results}
\label{sec:result}

We present a detailed evaluation of our trained policies in both 2-part assembly and multi-part assembly settings. 
The key takeaway is that our proposed approach can substantially improve individual policy performance, as well as the execution success of multi-step assembly sequences.

\subsection{Simulation-based Evaluation}
\label{subsec:sim-eval}

\subsubsection{BO-Guided Fine-Tuning}
\label{subsubsec:bo-finetuning}

We now present the results of our proposed approach that uses Bayesian Optimization (BO) to propose plug initialization samples for fine-tuning.

\begin{figure}
    \centering
    \includegraphics[width=\linewidth]{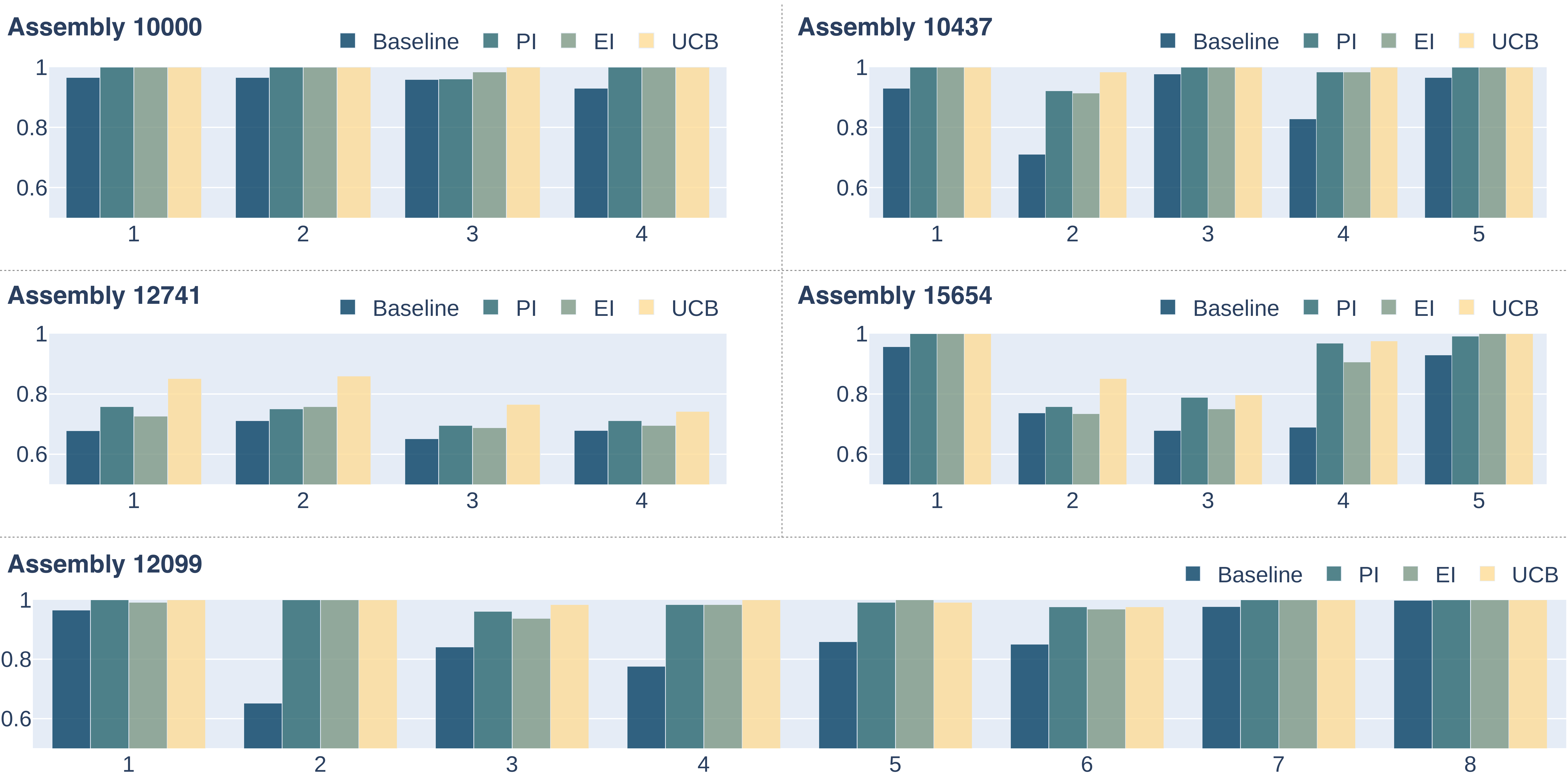}
    \caption{Comparison of success rate by using different acquisition function for fine-tuning in 5 multi-part assemblies (see \autoref{fig:multipart_dataset} for asset IDs). X-axis corresponds to step index within each assembly.}
    \label{fig:refinery_ac_comparison_per_assembly}
\end{figure}

Our first evaluation question is, \textbf{which acquisition function is the most effective for proposing plug initialization samples during fine-tuning for multi-part assembly?}
We evaluate our 3 acquisition functions on all 5 multi-part assemblies, which consist of 26 total assembly steps.
For each assembly step in each multi-part assembly, we train a specialist policy from scratch over 5 random seeds using \cite{tang2024automate} until convergence, and we use these policies as our baseline.
All baselines converge within 200 epochs, beyond which performance does not improve.
Then, for each acquisition function, we follow the procedure described in Algorithm \ref{algo:finetune} to fine-tune all 5 seeds of each specialist policy.
The fine-tuning procedure requires only 50 epochs, which is significantly fewer than training-from-scratch.
We evaluate each seed for 1000 trials and report the average success rate for each policy over the 5 seeds (\autoref{fig:refinery_ac_comparison_per_assembly}).
The average success rate is 83.61$\pm$9.27\% for baseline policies, 91.82$\pm$10.05\% for policies fine-tuned with PI as acquisition function, 91.12$\pm$10.61\% for policies fine-tuned with EI, and 94.41$\pm$7.51\% for policies fine-tuned with UCB.

\begin{figure}
    \centering
    \includegraphics[width=\linewidth]{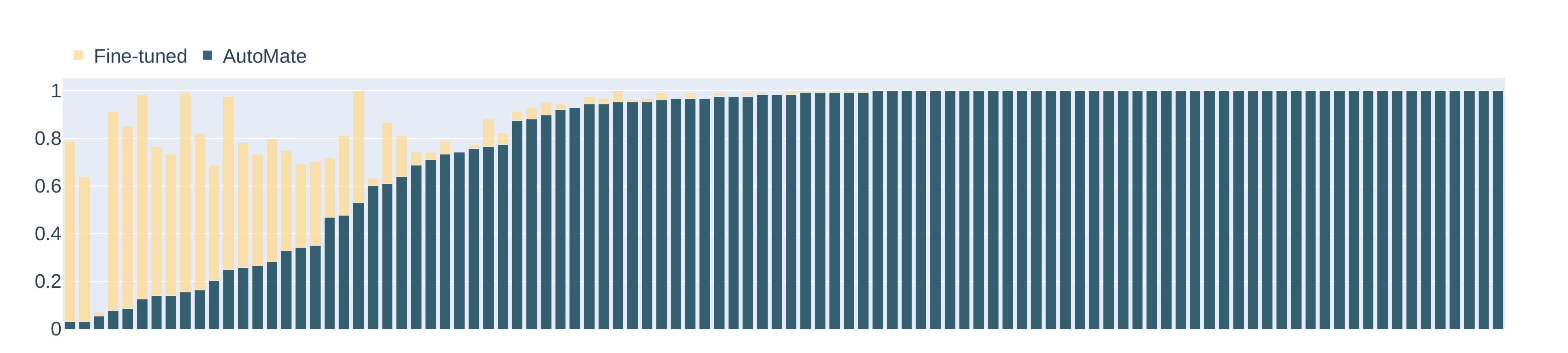}
    \caption{Average success rates of baseline policies (AutoMate) and fine-tuned policies across 100 2-part assembly tasks. Policies are sorted by baseline success rate. BO-guided fine-tuning significantly improves the average success rate from 80.53\% to 91.51\%, with greatest improvements occurring for the most challenging tasks.}
    \label{fig:sim-eval-100}
\end{figure}

Our second evaluation question is, \textbf{does our proposed fine-tuning approach also significantly improve policy performance compared to the state-of-the-art for 2-part assembly?}
We now use UCB as acquisition function and again follow the procedure described in Algorithm \ref{algo:finetune} to fine-tune specialist policies, but now over 100 2-part assembly tasks from an established baseline \cite{tang2024automate} \footnote{Visualization of the dataset can be found on their \href{https://bingjietang718.github.io/automate/}{project website}.}. 
We evaluate using the same procedure as for the prior evaluation.
The baseline policies, trained using \cite{tang2024automate} without any fine-tuning, achieve an average success rate of 80.53$\pm$31.43\%. After applying our proposed fine-tuning procedure, the average success rate increases to 91.51$\pm$14.27\%, demonstrating a clear performance gain (10.98\% increase in mean success rate and 17.16\% decrease in success rate variance). Improvement is consistently observed across the majority of tasks, with a number of fine-tuned policies approaching or reaching a 100\% success rate (\autoref{fig:sim-eval-100}).

\subsubsection{GMM-based Sampling for Deployment-Time Optimization}
\label{subsec:sim-eval-deployment}

We now present the results of our proposed approach that uses GMM-based sampling to determine plug initializations during policy deployment, as opposed to uniform sampling from the initialization space. Our evaluation question is, \textbf{does our proposed deployment-time optimization approach significantly improve performance of baseline and fine-tuned policies for multi-part assembly?}

We compare the following 4 approaches for training and deploying individual policies: (1) \textit{Baseline:} Train policy from scratch using \cite{tang2024automate} and uniformly sample plug initialization, (2) \textit{Deployment:} Train policy from scratch using \cite{tang2024automate} and sample plug initialization with GMM, (3) \textit{Fine-tune:} Fine-tune baseline policy with BO and uniformly sample plug initialization, (4) \textbf{\textit{Refinery}} (Ours): Fine-tune baseline policy with BO and sample plug initialization with GMM.

\begin{figure}
    \centering
    \includegraphics[width=\linewidth]{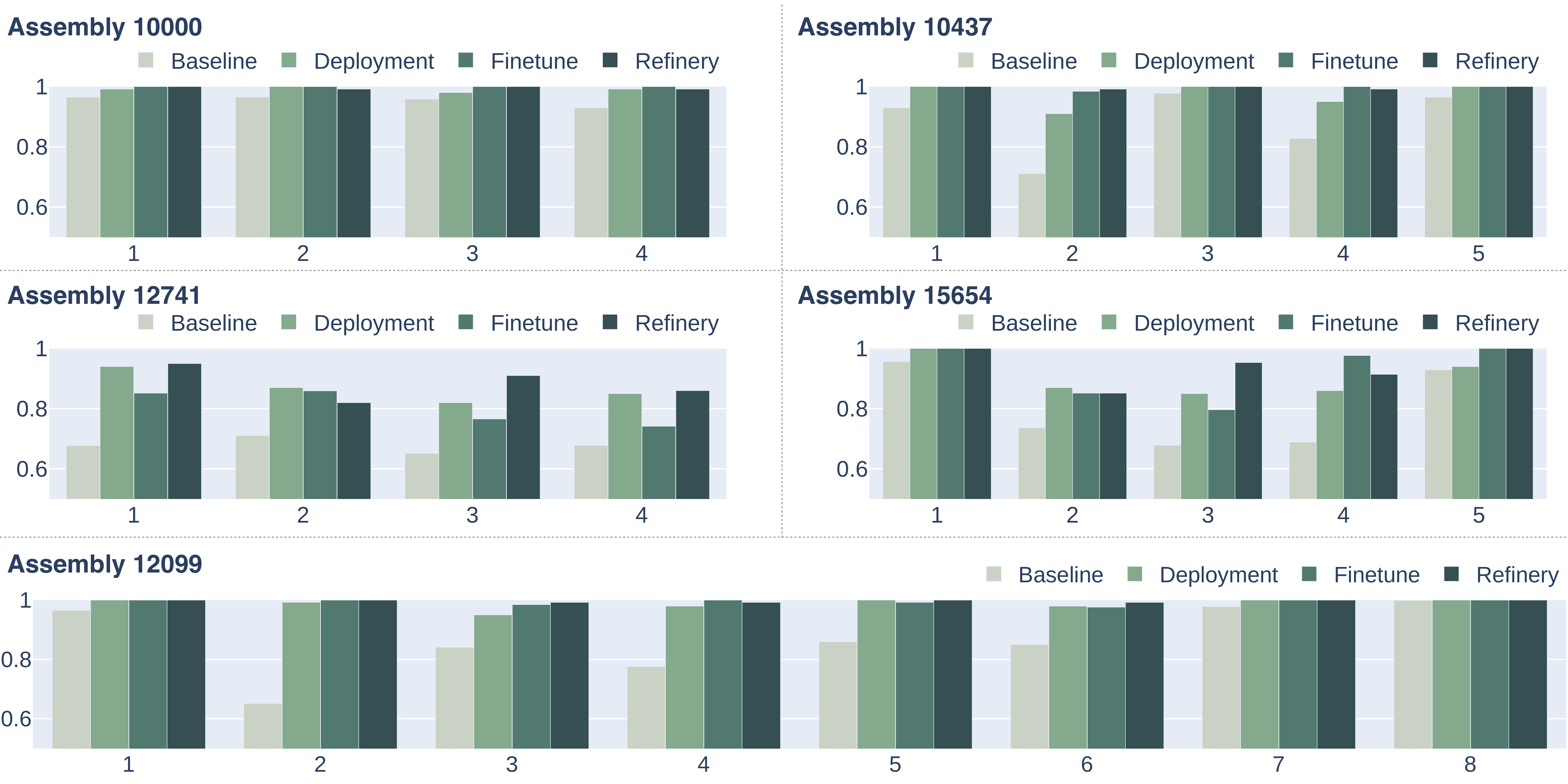}
    \caption{Comparison of success rate under different fine-tuning and deployment-time sampling strategies in 5 multi-part assemblies.}
    \label{fig:refinery_all_var_per_assembly}
\end{figure}

The policies are derived from Section~\ref{subsubsec:bo-finetuning}, which produced trained-from-scratch and UCB-fine-tuned policies for each assembly step over 5 random seeds.
For each of the 4 approaches listed above, we evaluate all 5 seeds for every assembly step over 1000 trials and calculate the average success rate (\autoref{fig:refinery_all_var_per_assembly}).
The initializations of all previously assembled parts are randomly sampled (i.e., the goal configuration is randomized), and we apply uniformly-sampled observation noise (within $\pm 2$mm on position, $\pm 10^{\circ}$ on orientation) on goal poses during training and evaluation.

The average success rate is 83.61$\pm$9.27\% for \textit{Baseline},
94.41$\pm$7.51\% for \textit{Fine-tune},
94.50$\pm$4.90\% for \textit{Deployment},
and 96.35$\pm$4.45\% for \textit{\textbf{Refinery}}.
Specifically, we observe that GMM-based initialization alone (\textit{Deployment}) improves average success rates over \textit{Baseline} without requiring additional online adaptation. 
This highlights the effectiveness of modeling success rate distributions using GMM and exploiting them for informed sampling during policy execution.
However, combining BO-guided fine-tuning with GMM-based deployment consistently achieves the highest success rates across all assemblies, outperforming both individual components. 
This demonstrates that while GMM-based sampling during deployment reduces the reliance on online fine-tuning, it does not eliminate the benefits of fine-tuning.
These results collectively suggest that to fully maximize success rates (critical for industrial settings), both a BO-based fine-tuning approach and GMM-based sampling strategy should be employed.

\subsubsection{Full Assembly Sequence}
\label{subsubsec:eval-full-assembly}
\begin{table}
  \centering
  \begin{tabular}{l|c|c|c|c}
\toprule
Asset ID & Baseline & Deployment & Fine-tune & Refinery (Ours) \\
\midrule
10000  & 83.20 & 96.47 {\scriptsize (+13.27)} & \textbf{99.22 {\scriptsize (+16.02)}} & 98.45 {\scriptsize (+15.25)} \\
12099  & 31.26 & 91.55 {\scriptsize (+60.29)} & 91.52 {\scriptsize (+60.26)} & \textbf{97.68 {\scriptsize (+66.42)}} \\
12741  & 21.10 & 56.78 {\scriptsize (+35.68)} & 48.65 {\scriptsize (+27.55)} & \textbf{60.62 {\scriptsize (+39.52)}} \\
10437  & 49.11 & 86.45 {\scriptsize (+37.34)} & 98.44 {\scriptsize (+49.33)} & \textbf{98.44 {\scriptsize (+49.33)}} \\
15654  & 31.98 & 56.48 {\scriptsize (+24.50)} & 66.22 {\scriptsize (+34.24)} & \textbf{74.19 {\scriptsize (+42.21)}} \\
\bottomrule
\end{tabular}
\caption{ Success rates for full assembly sequences.}
\label{tab:complete-assembly-success}
\end{table}

To demonstrate the effectiveness of the proposed combination of BO-guided fine-tuning and GMM-based sampling during deployment, we evaluate the success rate of the full assembly sequence across all 5 multi-part assemblies (\autoref{tab:complete-assembly-success}). 
In these experiments, a sequence is only considered as successful if all steps are executed successfully.
Our proposed method consistently outperforms the baseline across all evaluated assemblies, demonstrating its general applicability and robustness. 
Notably, assemblies with lower baseline performance, such as 12099 (31.26\%) and 15654 (31.98\%), exhibit substantial improvements under our approach, achieving success rates of 97.68\% and 74.19\%, respectively. 
Even in assemblies with relatively high baseline performance, such as 10000, our method yields a meaningful absolute gain of 15.25\%.
Overall, these findings underscore the synergistic benefit of combining uncertainty-aware fine-tuning with success-driven sampling for improving sequential task performance.

\subsection{Real-world Evaluation}
\label{subsec:real-eval}

\begin{figure}
    \centering
    \includegraphics[width=\linewidth]{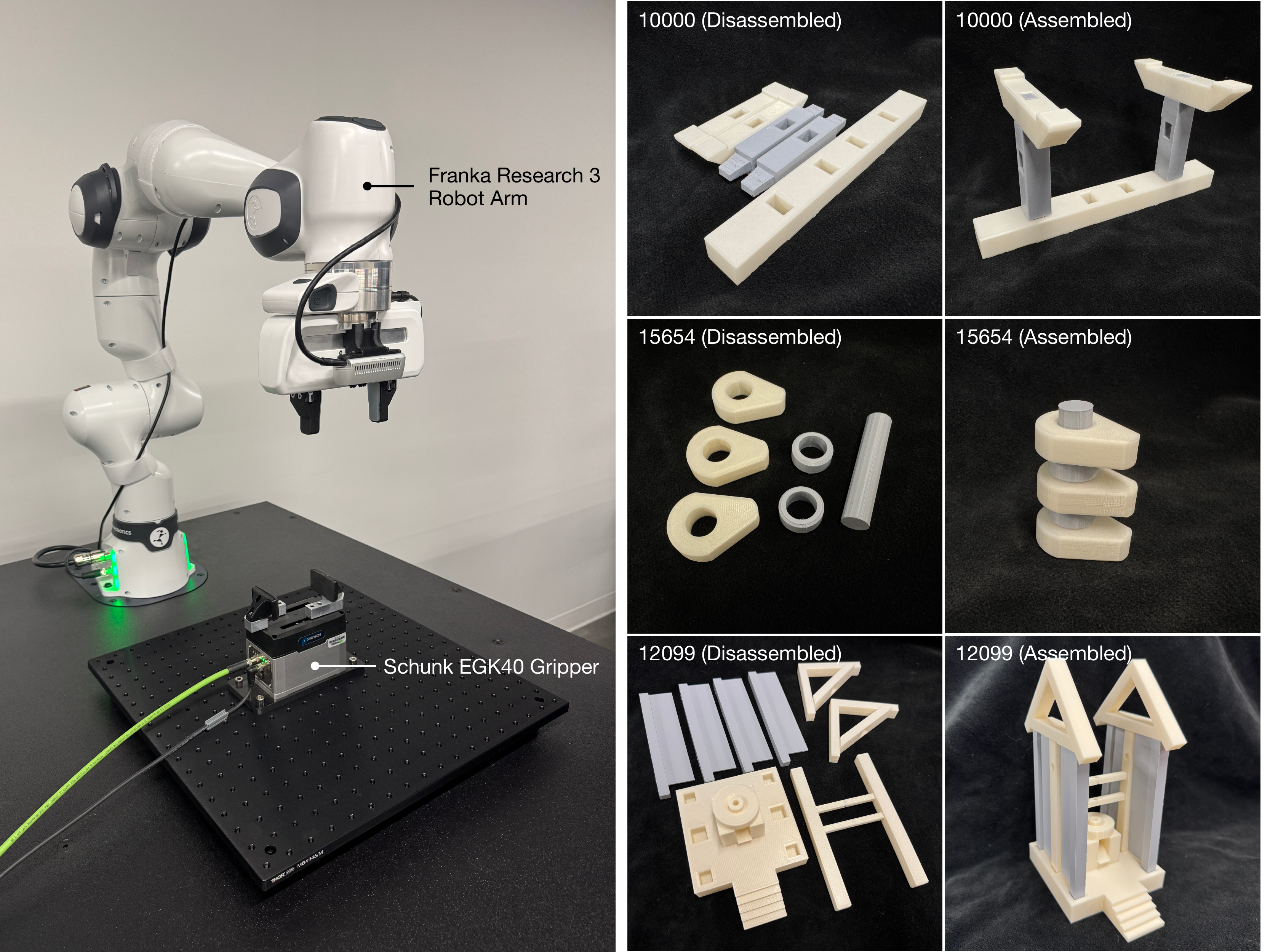}
    \caption{Real-world experiment setup. The Franka Research 3 (FR3) robot and a table-mounted Schunk EGK40 gripper is used to perform multi-part assembly tasks. 3D-printed 10000, 12099, and 15654 are shown in disassembled and assembled configurations.}
    \label{fig:real_exp_setup}
\end{figure}

To assess the real-world effectiveness of our approach, we evaluate performance across a diverse set of two-part and multi-part assemblies in the real world. Our real-world system (\autoref{fig:real_exp_setup}) consists of a robot arm with a parallel-jaw gripper, another parallel-jaw gripper mounted to the tabletop, and 3D-printed assemblies from our dataset (\autoref{fig:multipart_dataset}). 

\begin{table*}[]
\centering
\begin{tabular}{l|cc|cc|cc}
\toprule
& \multicolumn{2}{c|}{\textbf{Baseline}} & \multicolumn{2}{c|}{\textbf{Fine-tune}} & \multicolumn{2}{c}{\textbf{Refinery (Ours)}}\\
Asset ID & Simulation {\scriptsize (\%)} & Reality & Simulation {\scriptsize (\%)} & Reality & Simulation {\scriptsize (\%)} & Reality\\
\midrule
01053 & 56.25 & 2/10 & 91.41 & 7/10 & 97.66 & 9/10 \\
00190 & 54.69 & 4/10 & 87.34 & 10/10 & 99.22 & 10/10 \\
00514 & 76.56 & 6/10 & 88.28 & 8/10 & 89.84 & 8/10 \\
00614 & 84.37 & 8/10 & 89.69& 9/10 & 91.86 & 10/10 \\
00681 & 94.53 & 8/10 & 96.87 & 8/10 & 100.00 & 10/10 \\
00553 & 92.18 & 9/10 & 94.53 & 10/10 & 97.66 & 10/10 \\
00768 & 96.87 & 9/10 & 98.44 & 10/10 & 99.22 & 10/10 \\
00346 & 97.66 & 10/10 & 100.00 & 10/10 & 100.00 & 10/10 \\
01036 & 100.00 & 10/10 & 100.00 & 10/10 & 100.00 & 10/10 \\
01129 & 96.88 & 10/10 & 100.00 & 10/10 & 100.00 & 10/10 \\\midrule
Total & 84.99 & 76/100 & 94.66 {\scriptsize (+9.67)} & 92/100 {\scriptsize (+16)} & 97.54 {\scriptsize (+12.55)} & 97/100 {\scriptsize (+21)} \\
\bottomrule
\end{tabular}
\caption{Real world evaluation. Comparison of policy success rates under different fine-tuning and deployment-time sampling strategies on 10 two-part assemblies, over a total of 300 real-world trials.}
\label{tab:real-eval-automate}
\end{table*}

Our first evaluation question is, \textbf{does the proposed approach for fine-tuning and deployment also improve real-world policy performance?} We evaluate the \textit{Baseline, Fine-tune, and \textbf{Refinery}} approaches listed in \autoref{subsec:sim-eval-deployment} on 10 two-part assemblies. In these experiments, only \textit{\textbf{Refinery}} uses GMM-based initialization during deployment, and the GMM is directly modeled from data collected in simulation. For each of 10 assemblies, we deploy the corresponding policy for each approach 10 times, for a total of 300 trials. Our results are shown in \autoref{tab:real-eval-automate}. For assemblies where baseline policies already achieve high success rates (e.g., 00346, 01036, 01129), our method maintains near-perfect performance. In cases with moderate baseline success (e.g., 00614, 00681, 00553, 00768), the approach effectively eliminates residual failure cases, reducing the total number of failures from 3/40 to 0/40. For more challenging assemblies with low baseline success rates (e.g., 01053, 00190, 00514), the improvement is substantial, increasing the real-world success count from 12/30 to 27/30. The results demonstrate that our proposed approach consistently improves real-world policy performance without additional adaptation. 

\begin{table}
  \centering
    \begin{tabular}{l|c|c|c|c}
    \toprule
    Asset ID & Step & Fully Auto. & 1 Intervention & 2 Interventions \\ \midrule
    10000 & 37/40 & 8/10 & 10/10 & 10/10\\ 
    12099 & 65/70 & 6/10 & 10/10 & 10/10\\
    15654 & 44/50 & 5/10 & 9/10 & 10/10 \\\bottomrule
    \end{tabular}
    \caption{Success rates for full assembly and individual assembly in the real world across 3 multi-part assemblies.}
    \label{tab:real-eval-full-sequence}
\end{table}

Our second evaluation question is, \textbf{can we complete full sequences of multi-part assembly in the real world?} We evaluate our approach on three multi-part assemblies executed by a physical robot. For each assembly, the robot sequentially executes the learned policy for each subtask. If an intermediate step fails, a human operator intervenes to correct the failure before continuing. Each assembly is evaluated over 10 full-sequence trials, corresponding to 160 individual subtask executions. We report fully autonomous success rates and the success rate with 1 or 2 human interventions from the same experiments. As shown in \autoref{tab:real-eval-full-sequence}, the average fully-autonomous success rate across individual subtasks is 90.6\% (146/160), indicating strong per-step reliability. However, execution of the entire assembly sequence is more challenging: for 10000, a notable 80\% success rate was achieved over the sequence; for 15654, though, a 50\% success was achieved, primarily due to a single failing step, which is reflected in the success rates with only 1 or 2 human interventions.

\subsection{Limitations}

Across both simulation and real-world experiments, we identified the key causes of remaining failure cases. One cause was grasp instability, where the parallel-jaw gripper slipped or introduced unintended rotations during part transport. In addition, in multi-step assembly, the robot could apply large forces during policy execution, which  destabilized or displaced previously assembled parts. Additionally, in real-world trials, failures arose from structural limitations of 3D-printed components, such as flexing or tolerance mismatches with the 3D mesh due to printer settings, which could not be corrected by the policy.
These observations suggest that addressing robustness in grasping, integrating force control, and ensuring part stability may further advance long-horizon assembly.

\section{Conclusion}
\label{sec:conclusion}

We presented \textbf{Refinery}, a framework for improving the fine-tuning and deployment of learned policies in multi-part robotic assembly. By identifying high-uncertainty initial states with Bayesian Optimization and leveraging Gaussian Mixture Models to prioritize robust initializations at deployment, Refinery enhances both individual policy performance and long-horizon execution success. Our experiments show substantial gains on two-part and multi-part assembly tasks, achieving state-of-the-art results in both simulation and the real world. Several promising directions emerge from this work. Expanding the framework to more geometrically-diverse assembly tasks would test its generality beyond the current benchmark set. Incorporating perception-driven initialization could enable more autonomous deployment in unstructured environments. Additionally, exploring joint optimization strategies across policy chains may further improve sequence-level reliability in long-horizon assembly. Finally, future research will investigate how to bridge the gap towards additional industrial requirements, such as stricter tolerances and time constraints.

\printbibliography
\end{document}